\RequirePackage{fix-cm}
\documentclass[twocolumn]{svjour3}          
\smartqed  
\usepackage{graphicx}

 \usepackage{mathptmx}      
\usepackage{amsmath,amssymb,latexsym}
\usepackage{algorithm}
\usepackage{algorithmic}
\usepackage{bbm}
\usepackage{bm}
\usepackage{url}

\newcommand{\g}{\, | \, }

 \journalname{Statistics and Computing}
\begin{document}

\title{A Split-Merge MCMC Algorithm for the Hierarchical Dirichlet
Process }


\author{Chong Wang  \and David M. Blei }


\institute{Chong Wang \at
              Princeton University, Computer Science Department, Princeton, NJ, 08540\\
              Tel.: +1-609-258-1797\\
              Fax:  +1-609-258-1771\\
              \email{chongw@cs.princeton.edu}           
           \and
          David M. Blei \at
              Princeton University, Computer Science Department, Princeton, NJ, 08540\\
              Tel.: +1-609-258-9907\\
              Fax:  +1-609-258-1771\\
              \email{blei@cs.princeton.edu}           
}

\date{Received: date / Accepted: date}

\maketitle

\begin{abstract}
The hierarchical Dirichlet process (HDP) has become an important
Bayesian nonparametric model for grouped data, such as document
collections. The HDP is used to construct a flexible
mixed-membership model where the number of components is determined
by the data. As for most Bayesian nonparametric models, exact
posterior inference is intractable---practitioners use Markov chain
Monte Carlo (MCMC) or variational inference. Inspired by the
split-merge MCMC algorithm for the Dirichlet process (DP) mixture
model, we describe a novel split-merge MCMC sampling algorithm for
posterior inference in the HDP. We study its properties on both
synthetic data and text corpora.  We find that split-merge MCMC for
the HDP can provide significant improvements over traditional Gibbs
sampling, and we give some understanding of the data properties that
give rise to larger improvements.
\end{abstract}

\section{INTRODUCTION}
\label{sec:intro}

The hierarchical Dirichlet process (HDP)~\cite{Teh:2007} has become an important
tool for the unsupervised data analysis of grouped data~\cite{Teh:2010a}, for
example, image retrieval and object recognition~\cite{Li-Jia:2010},
multi-population haplotype phasing~\cite{Xing:2006}, time series
modeling~\cite{Fox:2009} and Bayesian nonparametric topic
modeling~\cite{Teh:2007}.  Specially, topic modeling is the scenario when the
HDP is applied to document collections, and each document is considered to be a
group of observed words.\footnote{We focus on topic modeling in this paper. We
will use ``HDP'' and ``HDP topic model'' interchangeably.}  This is an extension
of latent Dirichlet allocation (LDA)~\cite{Blei:2003b} that allows a potentially
unbounded number of topics (i.e., mixture components).  Given a collection of
documents, posterior inference for the HDP determines the number of topics from
the data.

As for most Bayesian nonparametric models, however, exact posterior
inference is intractable.  Practitioners must resort to approximate
inference methods, such as Markov chain Monte Carlo (MCMC)
sampling~\cite{Teh:2007} and variational inference~\cite{Teh:2007b}.
The idea behind both of these methods is to form an approximate
posterior distribution over the latent variables that is used as a
proxy for the true posterior.

We will focus on MCMC sampling, where the approximate posterior is
formed as an empirical distribution of samples from a Markov chain
whose stationary distribution is the posterior of interest.  The
central MCMC algorithm for the HDP is an incremental Gibbs
sampler~\cite{Teh:2007,Geman:1984}, which may be slow to mix (i.e.,
the chain must be run for many iterations before reaching its
stationary distribution). For example in topic modeling---where each
word of each document is assigned to a topic---the incremental Gibbs
sampler only allows changing the topic status of one observed word at
a time.  This precludes large changes in the latent
structure.\footnote{In~\cite{Teh:2007}, the Gibbs sampling based on
  the Chinese restaurant franchise (CRF) representation does allow the
  possibility of changing the status of some words in a document
  together. However, as stated in~\cite{Teh:2007}, this is a prior
  clustering effect and does not have practical advantages.}  Our goal
is to improve Gibbs sampling for the HDP.

We develop and study a split-merge MCMC algorithm for the HDP.  Our approach is
inspired by the success of split-merge MCMC samplers for Dirichlet process (DP)
mixtures~\cite{Jain:2004,Dahl:2003}\footnote{Here we focus on the conjugate
models. The non-conjugate version of split-merge MCMC samplers for DP mixtures
is presented in~\cite{Jain:2007}.}  The DP mixture~\cite{Antoniak:1974} is an
``infinite clustering model'' where each observation is associated with a single
component.  (In contrast, the HDP is a mixed-membership model.)  In split-merge
inference for DP mixtures, the Gibbs sampler is embellished with split-merge
operations.  Two observations are picked at random.  If the observations are in
the same component then a \textit{split} is proposed: all the observations
associated with that component are divided into two new components.  If the
observations are in different components then a \textit{merge} is proposed: the
observations from the two components are placed in the same component.  Finally,
whether the resulting split or merged state is accepted is determined by
Metropolis-Hastings. As demonstrated in \cite{Jain:2004,Dahl:2003}, split-merge
MCMC is effective for DP mixtures when the mixture models have overlapping
clusters.

Our split-merge MCMC algorithm for the HDP is based on the Chinese
restaurant franchise (CRF) representation of a two-level
HDP~\cite{Teh:2007}, where ``customers'' are partitioned at the
group-level and ``dishes'' are partitioned at the top level.  In an
HDP topic model, the customer partition represents the per-document
partition of words; the top level partition represents the sharing of
topics between documents. The split-merge algorithm for HDPs operates
at the top level---thus, the assignment of subsets of documents across
the corpus may be split or merged with other subsets.  (The reason we
don't do split-merge operations for the lower level DP is detailed in
Section~\ref{sec:sm-hdp}.)  We first demonstrate our algorithm on
synthetic data, and then study its performance on three real-world
corpora.  We see that our split-merge MCMC algorithm can provide
significant improvements over traditional Gibbs sampling, and we give
some understanding of the data properties that give rise to larger
improvements.

\section{THE HDP TOPIC MODEL}
\label{sec:hdp}

The hierarchical Dirichlet process~\cite{Teh:2007} is a hierarchical
generalization of the Dirichlet process (DP) distribution on random
distributions~\cite{Ferguson:1973}. We will focus on a two-level HDP,
which can be used in an infinite capacity mixed-membership model.  In
a mixed-membership model, data are groups of observations, and each
exhibits a shared set of mixture components with different proportion.
We will further focus on text-based topic modeling, the HDP topic
model.  In this setting, the data are observed words from a vocabulary
grouped into documents; the mixture components are distributions over
terms called ``topics.''

In an HDP mixed-membership model, each group is associated with a draw
from a shared DP whose base distribution is also a draw from a DP,
\begin{align*}
    G_0 & \sim {\rm DP}(\gamma, H) \\
    G_j \g G_0 & \sim {\rm DP}(\alpha_0, G_0), \mbox{ for each $j$},
\end{align*}
where $j$ is a group index.  At the top level, the distribution $G_0$
is a draw from a DP with concentration parameter $\gamma$ and base
distribution $H$.  It is almost surely discrete, placing its mass on
atoms drawn independently from $H$~\cite{Ferguson:1973}.  At the
bottom level, this discrete distribution is used as the base
distribution for each per-group distribution $G_j$.  Though they may
be defined on a continuous space (e.g., the simplex), this ensures
that the per-group distributions $G_j$ share the same atoms as $G_0$.

In topic modeling, each group is a document of words and the atoms are
distributions over words (topics).  The base distribution $H$ is
usually chosen to be a symmetric Dirichlet over the vocabulary
simplex, i.e., the atoms $\bm \phi=(\phi_k)_{k=1}^\infty$ are drawn
independently $\phi_k \sim {\rm Dirichlet}(\eta)$.  To complete the
HDP topic model, we draw the $i$th word in the $j$th document $x_{ji}$
as follows,
\begin{align}\label{eq:word-emit}
    \theta_{ji} \sim G_j, ~ ~ ~ x_{ji} \sim {\rm Mult}(\theta_{ji}).
\end{align}
We will show how $\theta_{ji}$ is related to $\bm \phi$ in next
section. The clustering effect of the Dirichlet process ensures that
this yields a mixed-membership model, where the topics are shared
among documents but each document exhibits them with different
proportion.  Based on this important property, we now turn to an
alternative representation of the HDP.

\subsection{The Chinese Restaurant Franchise}
\label{sec:crf}

Consider a random distribution $G$ drawn from a DP and a set of
variables drawn from $G$.  Integrating out $G$, these variables
exhibit a \textit{clustering effect}---they can be grouped according
to which take on the same value~\cite{Ferguson:1973}.  The values
associated with each group are independent draws from the base
distribution. The distribution of the partition is a Chinese
restaurant process (CRP)~\cite{Aldous:1985}.

In the HDP, the two levels of DPs enforce two kinds of grouping of the
observations.  First, words within a document are grouped according to
those drawn from the same ``unique'' atom in $G_j$.  (By ``unique,''
we mean atoms drawn independently from $G_0$.) Second, the {\it
  grouped} words in each document are themselves again grouped
according to those associated with the same atom in $G_0$.  Note that
this corpus-level partitioning connects groups of words from different
documents.  (And, it may connect two groups of words from the same
document.)  These partitions---the grouping of words within a document
and the grouping of word-groups within the corpus---are each governed
by a CRP.

As a consequence of this construction, the atoms of $G_0$, i.e., the population
of topics, are shared by different documents.  And, because of the clustering of
words, each document individually may exhibit several of those topics.  This is
the key property of the HDP.

\begin{figure}[t]
\centering
\includegraphics[width=.9\columnwidth]{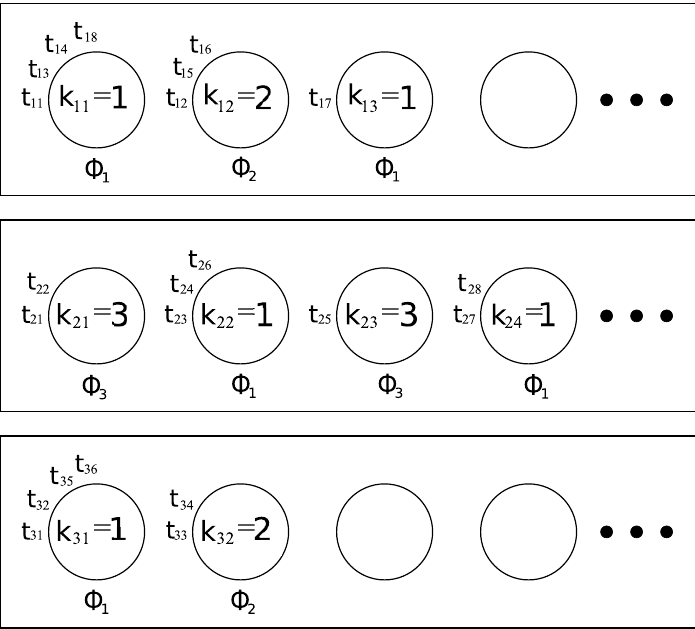}
\caption{A Depiction of a CRF, adapted from~\cite{Teh:2007}. Each
  restaurant/document is represented by a rectangle. Customer/word
  $x_{ji}$ is seated at a table (circles) in restaurant/document $j$
  via the customer-specific table index $t_{ji}$. Each table has a
  dish/topic, one of the global dishes/topics ($\phi_k$s), indicated
  by the table-specific dish/topic index, $k_{jt}$.}
\label{fig:crf}
\end{figure}


The split-merge Gibbs sampler that we develop below relies on a
representation of the HDP based on these partition probabilities,
rather than on the random distributions.\footnote{In the HDP in
  general, Gibbs samplers tend to operate on partitions; variational
  methods tend to operate on constructions of the random
  distributions.}  This representation is known as the Chinese
restaurant franchise (CRF)~\cite{Teh:2007}, a hierarchy of CRPs.  At
the document level, words in a document are grouped into ``tables''
according to a CRP for that document; at the corpus level, tables are
grouped into ``dishes'' according a corpus-level CRP.  All the words
that are attached (via their table) to the same dish are drawn from
the same topic.  This is illustrated in Figure~\ref{fig:crf}.


We now give the generative process for an HDP topic model based on the
CRF representation, which is important for developing the split-merge
MCMC algorithm.  Let $t_{ji}$ denote table index for $x_{ji}$, the
word $i$ in document $j$, and $k_{jt}$ denote the dish index, i.e.,
the global topic index, for table $t$ in document $j$.  The model
probabilities are based on several types of counts using on these
fundamental elements.  Notation is tabulated in Table~\ref{notation}.
\begin{table}
\begin{center}
\begin{tabular}{ll}
  \hline
  {\it notation} & {\it description} \\
  \hline
  $n_{jtk}$ &  \# words in document $j$ at table $t$ and topic $k$. \\
  $n_{jt\cdot}$ &  \# words in document $j$ at table $t$\\
  $n_{j\cdot k}$ & \# words in document $j$ belonging to topic $k$\\
  $n_{j\cdot \cdot}$ & \# words in document $j$\\
  $n_{\cdot \cdot k}$ & \# words belonging to topic $k$ in the corpus\\
  $m_{jk}$ &  \# tables in document $j$ belonging to topic $k$ \\
  $m_{j\cdot}$ &  \# tables in document $j$ \\
  $m_{\cdot k}$ &  \# tables belonging to topic $k$ in the corpus\\
  $m_{\cdot \cdot}$ & total tables in the corpus \\
  \hline
\end{tabular}
\end{center}
\caption{\label{notation}Notation used in the CRF representation}
\end{table}

There are three steps to the process.  First, we generate table
indices for each word in each document---this partitions the words
within the documents.  Then, we generate topic indices for each table
in each document---this partitions the tables (which are groups of
words) according to topics.  Finally, for each word we generate its
type (e.g., ``house'' or ``train'') from the assigned topic of its
assigned table.

\paragraph{Generate the table index $t_{ji}$.} In the document-level
CRP for document $j$, table index $t_{ji}$ is generated sequentially
according to
\begin{align}
p(t_{ji}=t \g t_{j 1}, \dots, t_{j,i-1}, \alpha_0)
\propto \left\{ \begin{array}{ll} n_{jt\cdot},&
    \textrm{if $t=1,\dots, m_{j\cdot}$}, \\ \alpha_0. & \textrm{if $t=t_{\rm new}$},
\end{array} \right. \nonumber
\end{align}
This induces the probability of partitioning the words of document $j$
into tables,
\begin{align}
    p(\bm t_j) = \frac{ \alpha_0^{m_{j\cdot}}
    \prod_{t=1}^{m_{j\cdot}} (n_{jt\cdot}-1)! } {\prod_{i=1}^{n_{j \cdot
    \cdot}}(i+\alpha_0-1)}.
\label{eq:doc-crp}
\end{align}

\paragraph{Generate the topic index $k_{jt}$.}
After all words in all documents are partitioned into tables, we
generate the topic index $k_{jt}$ for each table in each documents.
As above, the topic index comes from the corpus-level CRP,
\begin{align}\label{eq:crp-prob}
p(k_{jt} \g k_{11}, k_{12},  \dots, &\ k_{21}, k_{j,t-1}, \gamma) \\ 
&\propto \left\{ \begin{array}{ll} m_{\cdot k},&
    \textrm{if $k=1,\dots, K$}, \\ \gamma, & \textrm{if $k=k_{\rm new}$},
\end{array} \right. \nonumber
\end{align}
which induces the probability for the partition of all the tables.
Below, $D$ is the number of documents and $K$ is the number of used
topics in the set of topic indices,
\begin{align} \label{eq:corpus-crp}
    p(\bm k)= \frac{ \gamma^{K}\prod_{k=1}^{K}
    ((m_{\cdot k}-1)!} {\prod_{s=1}^{m_{\cdot \cdot}}(s+\gamma-1)}.
\end{align}

Let $\bm c$ denote the complete collection of table and topic indices,
$\bm t = (\bm t_j)_{j=1}^D$ and $\bm k$. We combine
Eq.~\ref{eq:doc-crp} and~\ref{eq:corpus-crp},
\begin{align}
    p({\bm c}) \textstyle = p(\bm k)p(\bm t) =  p(\bm k)\prod_{j=1}^D p(\bm t_j).
    \label{eq:partition-prob}
\end{align}
\paragraph{Generate word observations $x_{ji}$.}
Finally, we generate the observed words $x_{ji}$.  In the previous
representation of the HDP in Eq.~\ref{eq:word-emit}, the words were
generated given a topic $\theta_{ji}$.  In this representation, each
$x_{ji}$ is associated with a table index $t_{ji}$, and each table is
associated with a topic index $k_{jt}$, which links to one of the
topics $\phi_k$.  Define $z_{ji} = k_{jt_{ji}}$,
\begin{align*}
  \theta_{ji} = \phi_{z_{ji}}, ~ ~ ~ x_{ji} \sim {\rm
    Mult}(\theta_{ji}).
\end{align*}
We call $z_{ji}$ the topic index for word $x_{ji}$.
It locates the topic $\phi_k$ from which  $x_{ji}$ is generated.


Since different words with the same value of $z_{ji}$ are drawn from
the same topic $\phi_{z_{ji}}$, we can consider the conditional
likelihood of the corpus $\bm x = (\bm x_j)_{j=1}^D$ given all the
latent indices $\bm c$.  We are integrating out the topics $\bm \phi$.
(Recall $\phi_k \sim {\rm Dirichlet}(\eta)$.)  This conditional
likelihood is
\begin{align}
 p(\bm x|\bm c) = \textstyle \prod_k f_k(\{x_{ji}:z_{ji}=k\}), \label{eq:likelihood},
\end{align}
where
\begin{align*}
f_k(\{x_{ji}:z_{ji}=k\}) = \frac{\Gamma(V\eta)}{\Gamma(n_{\cdot \cdot k}+V\eta)}
\frac{\prod_v\Gamma(n_{\cdot \cdot k}^{v}+\eta)}{\Gamma^V(\eta)}.
\end{align*}
The size of the vocabulary is $V$ and the number of words assigned to
topic $k$ is $n_{\cdot \cdot k}^{v}$. This completes the generative
process for an HDP topic model.

\section{SPLIT-MERGE MCMC FOR THE HDP}
\label{sec:sm-hdp}

Given a collection of documents, the goal of posterior inference is to
compute the conditional distribution of the latent structure, the
assignment of documents to topics and the distributions over words
associated with each topic.  In~\cite{Teh:2007} posterior inference is
based on the CRF representation presented above.  Their Gibbs sampler
iteratively samples the table indices $\bm t$ and topic indices $\bm
k$. Details are in~\cite{Teh:2007}.

We now develop split-merge MCMC for the HDP.  Our motivation is that
incremental Gibbs samplers can be slow to converge, as they only
sample one variable at a time.  Split-merge algorithms can consider
larger moves in the state space and have the potential to converge
more quickly.  To construct the split-merge MCMC algorithm for the
HDP, we first recall that, from Eq.~\ref{eq:crp-prob}
and~\ref{eq:corpus-crp}, the top level of the HDP can be described as
the corpus-level CRP with tables from all documents as
observations. The idea behind our algorithm is to use split-merge MCMC
algorithm for the DP mixtures~\cite{Jain:2004,Dahl:2003} at this
top-level.

To make the above concrete, we first review the traditional Gibbs
sampling algorithm for the topic indices $\bm k$ and then present the
split-merge MCMC algorithm. At the end of this section, we discuss why
we only use split-merge operations on the top level of the HDP.

\subsection{Gibbs sampling for topic indices ${\bm k}$} 
\label{sec:tgibbs}

Since all tables in the corpus are partitioned into topics according
to the corpus-level CRP (see Eq.~\ref{eq:crp-prob}
and~\ref{eq:corpus-crp}), we follow the standard procedure (Algorithm
3 in Neal's paper~\cite{Neal:2000}) to derive their Gibbs sampling
updates.  Let $f_k^{-{\bm x}_{jt}}({\bm x}_{jt})$ denote the
conditional density of ${\bm x}_{jt}$ (all the words at table $t$ in
document $j$) given all words in topic $k$, excluding ${\bm x}_{jt}$,
\begin{align}
& f_k^{-{\bm x}_{jt}}({\bm x}_{jt}) = \label{eq:f-k} \\
&\frac{\Gamma(n_{\cdot \cdot
k}^{-{\bm x}_{jt}}+V\eta)}{\Gamma(n_{\cdot \cdot
k}^{-{\bm x}_{jt}}+n^{{\bm x}_{jt}}+V\eta)} \frac{\prod_v\Gamma(n_{\cdot \cdot
k}^{-{\bm x}_{jt},v}+n^{{\bm x}_{jt},v}+\eta)}{\prod_v\Gamma(n_{\cdot \cdot
k}^{-{\bm x}_{jt},v}+\eta)}, \nonumber
\end{align}
where $n_{\cdot \cdot k}^{-x_{ji},v}$ is the number of word $v$ with topic $k$,
excluding $x_{ji}$. See the appendix for the relevant derivations, which is same
as in~\cite{Teh:2007}.

Because of the exchangeability of the CRP, we can view the table
$k_{jt}$ as the last table. Thus, by combining Eq.~\ref{eq:crp-prob}
and~\ref{eq:f-k}, we obtain the Gibbs sampling algorithm for
$k_{jt}$~\cite{Neal:2000},
\begin{align*}
p(k_{jt}=k  \g{\bm t}, {\bm k}^{-jt}, \bm x) \propto \left\{ \begin{array}{ll} m_{\cdot k}^{-jt} f_{k}^{-{\bm x}_{jt}}({\bm x}_{jt}),&
      \textrm{if $k=1,., K$}, \\ \gamma f_{k_{\rm
          new}}^{-{\bm x}_{jt}}({\bm x}_{jt}), & \textrm{if $k=k_{\rm new}$}.
    \end{array} \right. \nonumber
\end{align*}
Note that changing $k_{jt}$ changes the topic of all the words in $\bm
x_{jt}$ together. However, as discussed in~\cite{Teh:2007}, assigning
words to different tables with the same $k$ is a {\it prior}
clustering effect of a DP with $n_{j\cdot k}$ customers (and only
happens in one document).  Reassigning $k_{jt}$ to other topics is
unlikely.

\subsection{A split-merge algorithm for topic indices $\bm k$}
\label{sec:sm-gibbs}

The split-merge MCMC algorithm for DP mixture
models~\cite{Jain:2004,Dahl:2003} starts by randomly choosing two
observations. If they are in the same component then a {\it split} is
proposed, where all the observations in this component are assigned
into two new components (and the old componenti is removed).  If they
are in two different components then a {\it merge} is proposed, where
all the observations in the two components are merged into one new
component (and the two old components are removed). Whether the
proposal of split or merge is accepted or not is determined by the
Metropolis-Hastings ratio.

Based on the discussion in section~\ref{sec:tgibbs}, the split-merge
MCMC algorithm for DP mixture models can be used at the top level of
the HDP by viewing the tables as the observations for the corpus-level
CRP.  We present our algorithm using the sequential allocation
approach proposed in~\cite{Dahl:2003} . (This is easy to modify to use
the intermediate scans used in~\cite{Jain:2004}.) Thus, the sketch of
this algorithm is similar to the one presented in~\cite{Dahl:2003}.

We describe the procedure when a {\it split} is proposed. Two tables
have been selected and their assigned topics are the same---this is
{\it the selected topic}. We then create two new topics with each
containing one of the tables just selected. Finally, we consider all
the other tables in the corpus assigned to the selected topic, and
assign those tables into the two new topics.
Following~\cite{Dahl:2003}, this is done by running a ``mini
one-pass'' Gibbs sampler over only the two new topics, and
partitioning each table into one or other. We call the new state
containing the two new topics the {\it split state}.

In more detail, let $\bm c$ be the current state and ${\bm c}_{\rm
  split}$ be the split state. We use $(j, t)$ to indicate the table
$t$ in document $j$.  Then two selected tables are represented as
$(j_1, t_1)$ and ($j_2,t_2$). In state $\bm c$, the selected topic is
$k=k_{j_1, t_1} =k_{j_2, t_2}$.  Further, let $S_{\bm c}$ be the set
of tables whose topic is $k$ excluding table $(j_1, t_1)$ and $(j_2,
t_2)$,
\[
S_{\bm c}=\{(j,t): (j,t)\neq (j_1, t_1), (j,t) \neq (j_2, t_2), k_{jt}=k \}.
\]
In state $\bm c_{\rm split}$, in order to replace topic $k$, we create
two new topics, $k_1$ and $k_2$, then assign $k_{j_1, t_1} = k_1$ and
$k_{j_2, t_2} = k_2$ for the two initially selected tables. Let us now
show how we use {\it sequential allocation restricted Gibbs sampling}
(the ``mini one-pass Gibbs sampler'' we just mentioned)~\cite{Dahl:2003}
to reach split state $\bm c_{\rm split}$ from state $\bm c$.

{\noindent \bf Sequential allocation restricted Gibbs sampling.}  Define
$S_1=\{(j_1, t_1)\}$ and $S_2=\{(j_2,t_2)\}$ and recall that $k_{j_1,
  t_1} = k_1$ and $k_{j_2, t_2} = k_2$. Let $m_{\cdot k_1}=|S_1|$ and
$m_{\cdot k_2}=|S_2|$ be the number of tables in $S_1$ and $S_2$.  We
will use set $S_1$ or $S_2$ to receive the tables from $S_{\bm c}$
that will be assigned to $k_1$ or $k_2$ in the sequential restricted
Gibbs sampling procedure. Let $(j, t)$ be successive table indexes in
a uniformly permuted $S_{\bm c}$ and sample $k_{jt}$ according to the
following,
\[
p(k_{jt} = k_{\ell}|S_1, S_2) \propto m_{\cdot k_{\ell}}f_{k_{\ell}}^{-\bm x_{jt}} (\bm
x_{jt} ), \ell =1, 2.
\]
This is a one-pass Gibbs sampling of $k_{jt}$ but restricted over only
topic $k_1$ and $k_2$--called {\it sequential allocation restricted
  Gibbs sampling} in~\cite{Dahl:2003}. We have
\begin{align*}
\mbox{If $k=k_1$, then } &S_1 \leftarrow S_1 \cup (j, t),  ~ ~ m_{\cdot k_1} \leftarrow   m_{\cdot k_1} + 1, \\
\mbox{otherwise, } &S_2 \leftarrow S_2 \cup (j, t), ~ ~  m_{\cdot k_2} \leftarrow   m_{\cdot k_2} + 1
\end{align*}
Repeat this until all tables in $S_{\bm c}$ are visited. The final
$S_1$ and $S_2$ will contain all the tables that are assigned to $k_1$
and $k_2$ in state $\bm c_{\rm split}$.  Let the realization of
$k_{jt}$ be $k_{jt}(r)$, then we can compute the transition
probability from state $\bm c$ to split state $\bm c_{\rm split}$ as
\[
q(\bm c\rightarrow {\bm c}_{\rm split}) = \textstyle \prod_{(j,t) \in S}  p(k_{jt} =
k_{jt}(r)|S_1, S_2).
\]
Note that we have abused notation slightly, since $S_1$ and $S_2$ are
changing when we sample $k_{jt}$. The transition probability from
split state $\bm c_{\rm split}$ to state $\bm c$ is
\[
q(\bm c_{\rm split} \rightarrow {\bm c})=1.
\]
This is because there is only one way to merge two topics $k_1$ and $k_2$ in
state $\bm c_{\rm split}$ into topic $k$ in state $\bm c$.


Now we decide whether to retain the new partition, with the split
topics in place of the original topic, or whether to return to the
partition before the split. This decision is sampled from the
Metropolis-Hastings acceptance ratio.  The probability of accepting
the split is
\[
\mathcal{A}=\frac{p({\bm c}_{\rm split}) } {p({\bm c})} \frac { L({\bm c}_{\rm split})
}{ L({\bm c})} \frac{q(\bm c_{\rm split} \rightarrow {\bm c})} {q(\bm c
\rightarrow{\bm c}_{\rm split})}.
\]
We obtain the prior ratio from Eq.~\ref{eq:partition-prob},
\[
\frac{p({\bm c}_{\rm split})} {p({\bm c})} = \gamma \frac{ (m_{\cdot k_1}-1)!
  (m_{\cdot k_2}-1)!} {(m_{\cdot k}-1)!}.
\]
Define $L(\bm c)=p(\bm x\g \bm c)$ in Eq.~\ref{eq:likelihood}.  The
likelihood ratio is
\[
\frac{L({\bm c}_{\rm split})} {L({\bm c})} =
\frac{f_{k_1}(\{x_{ji}:z_{ji}=k_1)\}f_{k_2}(\{x_{ji}:z_{ji}=k_2\})}
{f_{k}(\{x_{ji}:z_{ji}=k\})}.
\]
The validity of the MCMC proposal can be verified by
following~\cite{Jain:2004,Dahl:2003}. The complete algorithm is
described in Figure~\ref{alg:sm-mcmc}, which also contains the merge
proposal, i.e., the case where the two initially selected tables are
attached to different topics.

\begin{figure*}[t]
\hrule
Assume the current state is $\bm c$. 
\begin{enumerate}
\item Choose two distinct tables, $(j_1, t_1)$ and  $(j_2, t_2)$, at random uniformly.  
\item {\bf Split case}: if {$k_{j_1t_1} = k_{j_2t_2}=k$,} then
  \begin{enumerate}

      \item Let $S_{\bm c}$ the set of tables whose topic is $k$ excluding table
          $(j_1, t_1)$ and $(j_2, t_2)$ in state $\bm c$, that is $S_{\bm
          c}=\{(j,t): (j,t)\neq (j_1, t_1), (j,t) \neq (j_2, t_2), k_{jt}=k \}$.

  \item Assign $k_{j_1t_1}=k_1$ and $k_{j_2t_2}=k_2$. Randomly permute $S_{\bm
      c}$, then run sequential allocation restricted Gibbs algorithm to assign
      the tables in $S_{\bm c}$ to $k_1$ or $k_2$ to obtain the split state $\bm
      c_{\rm split}$.  Calculate the product of the probability used as $q(\bm
      c\rightarrow {\bm c}_{\rm split})$.
  \item Calculate the acceptance ratio:
    \[
    \mathcal{A}=\frac{p({\bm c}_{\rm split}) L({\bm c}_{\rm split})} {p({\bm c})
      L({\bm c})} \frac{q({\bm c}_{\rm split} \rightarrow {\bm c})} {q(\bm c
      \rightarrow  {\bm c}_{\rm split})},
    \]
    where $q(\bm c_{\rm split} \rightarrow \bm c)=1$, since there is
    only one way to merge two topics from ${\bm c}_{\rm split}$ to ${\bm
      c}$.
  \end{enumerate}
\item {\bf Merge case}: if $(k_{j_1t_1}=k_1) \neq (k_{j_2t_2}=k_2)$,
  \begin{enumerate}
      \item Let $S_{\bm c}$ be the set of tables whose topics are either $k_1$
          or $k_2$ excluding table $(j_1, t_1)$ and $(j_2, t_2)$ in state $\bm
          c$, that is $S_{\bm c}=\{(j,t): (j,t)\neq (j_1, t_1), (j,t) \neq (j_2,
          t_2), k_{jt}=k_1 \mbox{ or } k_{jt}=k_2 \}$.

      \item  Randomly permute $S_{\bm c}$, then run sequential allocation
          restricted Gibbs algorithm to assign the tables in $S_{\bm c}$ to
          $k_1$ or $k_2$ to reach the original split state $\bm c$. Calculate
          the product of the probability used as $q(\bm c_{\rm merge}
          \rightarrow \bm c)$.

      \item Assign  $k_{j_1t_1}=k_{j_2t_2}=k$ and $k_{jt}=k$ for $(j,t) \in
          S_{\bm c}$ to obtain merge state $\bm c_{\rm merge}$.

     \item {Calculate the acceptance ratio:
        \[
        \mathcal{A}=\frac{p({\bm c}_{\rm merge}) L({\bm c}_{\rm merge})} {p({\bm c})
          L({\bm c})} \frac{q(\bm c_{\rm merge} \rightarrow {\bm c})} {q(\bm c\rightarrow
          {\bm c}_{\rm merge})},
        \]
        where $q(\bm c \rightarrow {\bm c}_{\rm merge})=1$, since there is only one way to merge two
        topics from $\bm c$ to ${\bm c}_{\rm merge}$. }
    \end{enumerate}
  \item {Sample $u\sim {\rm Unif(0,1)}$, if $u<\mathcal{A}$, accept the move;
      otherwise, reject it.}
  \end{enumerate}
\hrule
\caption{The split-merge MCMC algorithm for the HDP.}
\label{alg:sm-mcmc}
\end{figure*}


{\noindent \bf Discussion.} There are two other possibile ways to introduce split-merge
operations in the HDP.  First, we can split and merge within the document-level
DP.  This is of little interest because we care more about the words global
topic assignments than their local table assignments.  Second, we can split and
merge by choosing two words and resample all the other words in the same
topic(s).  However, this will be inefficient. Two different words have zero
similarity, and so two random picked words will hardly serve a good guidance and
could have a very low acceptance ratio.  (This is different from a DP mixture
model applied to continuous data, where different data points can have different
similarities/distances.).  In contrast, tables can be seen as word vectors with
many non-zero entries, which mitigates this issue.


\section{EXPERIMENTS}
\label{sec:exp}

We studied split-merge MCMC for the HDP topic model on synthetic and real data.
To initialize the sampler, we use sequential prediction---we iteratively assign
words to a table and a topic according to the predictive distribution given the
previously seen data and the algorithm proceeds until all words are``added''
into the model.  This works well empirically and was used in~\cite{Liang:2007a}
for DP mixture models. In addition, we use multiple random starts.  Our C++ code
is available as a general software tool for fitting HDP topic models with
split-merge and traditional Gibbs algorithms at \\
\url{http://www.cs.princeton.edu/~chongw/software/hdp.tar.gz}.


\subsection{Synthetic Data}
\label{sec:syn-data}

\begin{figure}[t]
\centering
\includegraphics[width=1\columnwidth]{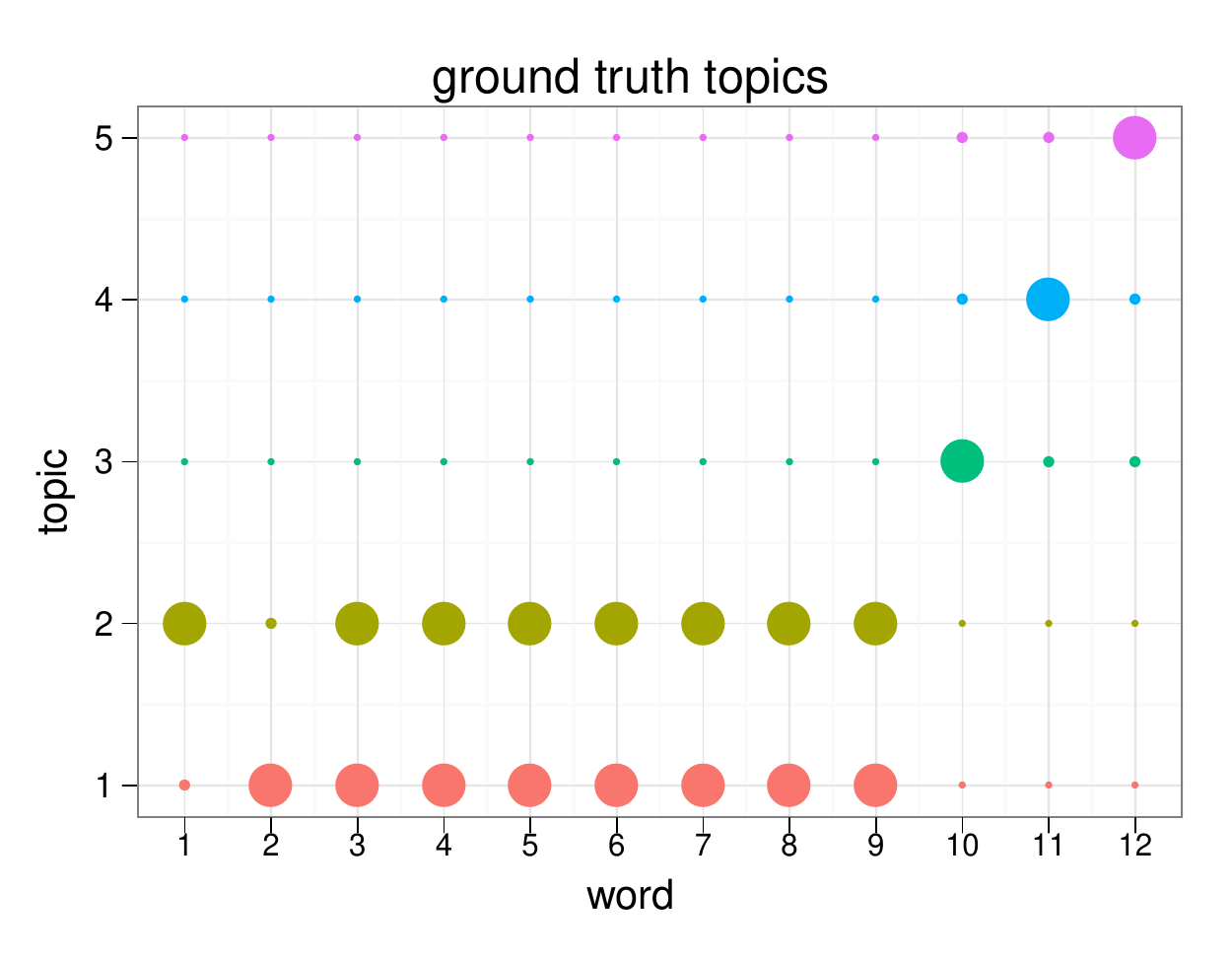}
\caption{The ``word'' and ``topic'' axes indicate the word and topic
  indexes.  The different sizes of the dots indicate the relative
  probability values of the words in that topic.}
\label{fig:groundtruth}
\end{figure}

We use synthetic text data to give an understanding of how the
algorithm works.  We generated 100 documents, each with 50 words, from
a model with 5 topics.  There are 12 words in the vocabulary and each
document uses at most 2 topics.  The topic multinomial distributions
over the words are shown in Figure~\ref{fig:groundtruth}.
In this model, topics 1 and 2 are very similar---they share 7 words
(word 3 to 9) with the same highest probability.  Topic 1 places high
probability on word 1 but low probability on word 2; topic 2 is
reversed. Others topics are less similar to each other---topics 3-5
share no words with 1 and 2 and have different distributions over the
remaining words.  Thus, it's expected that topics 1 and 2 are
difficult to distinguish; identifying the rest should be easier.  Our
goal is to demonstrate that without split-merge operations, it is
difficult for the traditional Gibbs sampler to separate topics 1 and
2.

For the HDP topic models, we set the topic Dirichlet parameter $\eta =
0.5$, hyperparameters $\gamma$ and $\alpha$ with Gamma priors ${\rm
  Gamma}(0.1, 1)$ to favor sparsity. (Without sampling $\gamma$ and
$\alpha$, the results are similar.) For this experiment, we run one
split-merge trial after each Gibbs sweep.  We run the algorithms for
1000 iterations.

\begin{figure*}[t]
\centering
\includegraphics[width=1\textwidth]{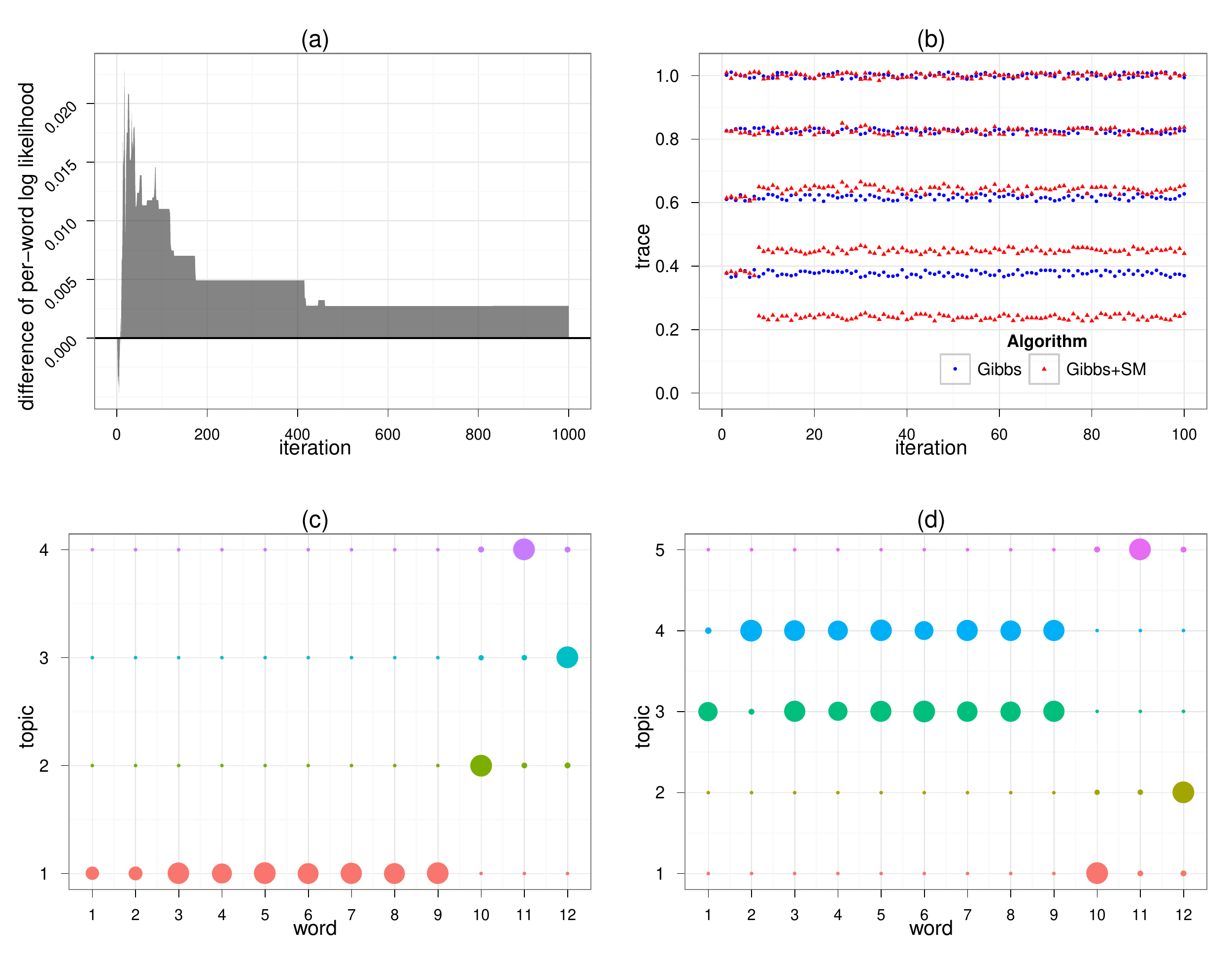}
\caption{Experimental results on synthetic data.  (best viewed in color.) The
new algorithm is ``Gibbs+SM''.  (a) The difference of the best per-word log
likelihood up to the same time, where above zero indicates that Gibbs+SM does
better. (b) Trace plot on the ratios truncated at the 100 iteration. The rest of
iterations is similar. Points are jittered for better view.  (c) Topic
visualization for ``Gibbs''on its best mode. There are 4 topics indicated along
the ``topic'' axis, also with different colors. (d) Topic visualization for
``Gibbs+SM'' at its best mode.  There are 5 topics.}
\label{fig:synthetic}
\end{figure*}

Figure~\ref{fig:synthetic} shows the results. In
Figure~\ref{fig:synthetic}(a), we compare the {\it modes} by plotting
the difference of the {\it best} per-word log likelihood up to the
{same} time (here time is the same as iteration), which is,
\begin{align}
  y_t = M_{t, Gibbs+SM} - M_{t, Gibbs},
\end{align}
where $M_{t, Gibbs+SM}$ and $M_{t,Gibbs}$ indicate the modes found
before time $t$, i.e., the best per-word log likelihoods for Gibbs
sampling with split-merge and the pure Gibbs sampling up to time
$t$. (This log likelihood is proportional in log space to the true
posterior---higher log likelihood indicates a state with higher
posterior probability.)  We found that split-merge explores the space
to a better mode.


In Figure~\ref{fig:synthetic}(b), we compared the topic trace plot,
which contains cumulative ratios of the words assigned to the most
popular, two most popular, \dots, to all topics.  (This was adapted
from~\cite{Jain:2004}.)  Ideally, for our problem, these will be 0.2,
0.4, 0.6, 0.8 and 1.0. In this experiment, the traditional Gibbs
sampler gets trapped with 4 topics while our algorithm finds 5 topics
after several iterations.

In Figure~\ref{fig:synthetic}(c) and (d), we visualized the topics
obtained by each algorithm. Both methods identify the three easy
topics---topics 3, 4 and 5 in the data, and these correspond to topic
2, 3 and 4 in Figure~\ref{fig:synthetic}(c) and topic 1, 2 and 5 in
Figure~\ref{fig:synthetic}(d).  However, the traditional Gibbs
sampling cannot identify the difference between topic 1 and 2 in the
data---see topic 1 in Figure~\ref{fig:synthetic}(c), which is a
combination of the two true topics.  In contrast, the split-merge
algorithm distinguishes them---see topic 3 and 4 in
Figure~\ref{fig:synthetic}(d).

Split-merge algorithm introduces new Metropolis-Hastings moves and,
thus, is computationally more expensive than the traditional Gibbs
sampling.  However, we only split or merge at the top-level DP, where
a table is treated as an observation, and the number of tables is
usually much smaller than the number of words.  So, we expect the
additional expense to be minimal.  In the synthetic data, the
difference was negligible.

Finally, in the experiments in~\cite{Jain:2004} for DP mixtures,
reverse split-merge moves (i.e. from state $A$ to state $B$, then from
state $B$ to state $A$) are frequent, while we do not see this
behavior here.  We hypothesize that this is because large moves, like
a split or merge, are only accepted when the HDP reaches a much better
local mode and therefore the chance of making the reverse move is very
small.  Although running the Markov chain for a sufficient long time
might mitigate this issue, in practice we recommend running the
split-merge operations in the burn-in phase.  (This is also the
strategy we use in the analysis on real data.)

\subsection{Analysis of Text Corpora}
\label{sec:real-data}

We studied split/merge MCMC on three text corpora:
\begin{itemize}
\item {\it ARXIV}: This is a collection of 2000 abstracts (randomly sampled)
    from online research abstracts\footnote{\url{http://arxiv.org}}. The vocabulary has 2441
    unique terms and the entire corpus contains around 89K words.
\item {\it ML+IR}: This is a collection of 2080 conference abstracts downloaded
    from machine learning (ML) and informational retrieval (IR) conferences,
    including CIKM, ICML, KDD, NIPS, SIGIR and WWW from year
    2005-2008~\cite{Wang:2009}\footnote{\url{http://www.cs.princeton.edu/~chongw/data/6conf.tgz}}.
    The vocabulary has 3237 unique terms and the corpus contains around 118K
    words.
\item {\it NIPS}: This is a collection of 1392 abstracts, a subset of the NIPS
    articles published between
    1988-1999\footnote{\url{http://www.cs.utoronto.ca/~sroweis/nips}}.  The
    vocabulary has 4368 unique terms and the entire corpus contains around 263K
    words.
\end{itemize}

\begin{figure}[t]
\centering
\includegraphics[width=0.95\columnwidth]{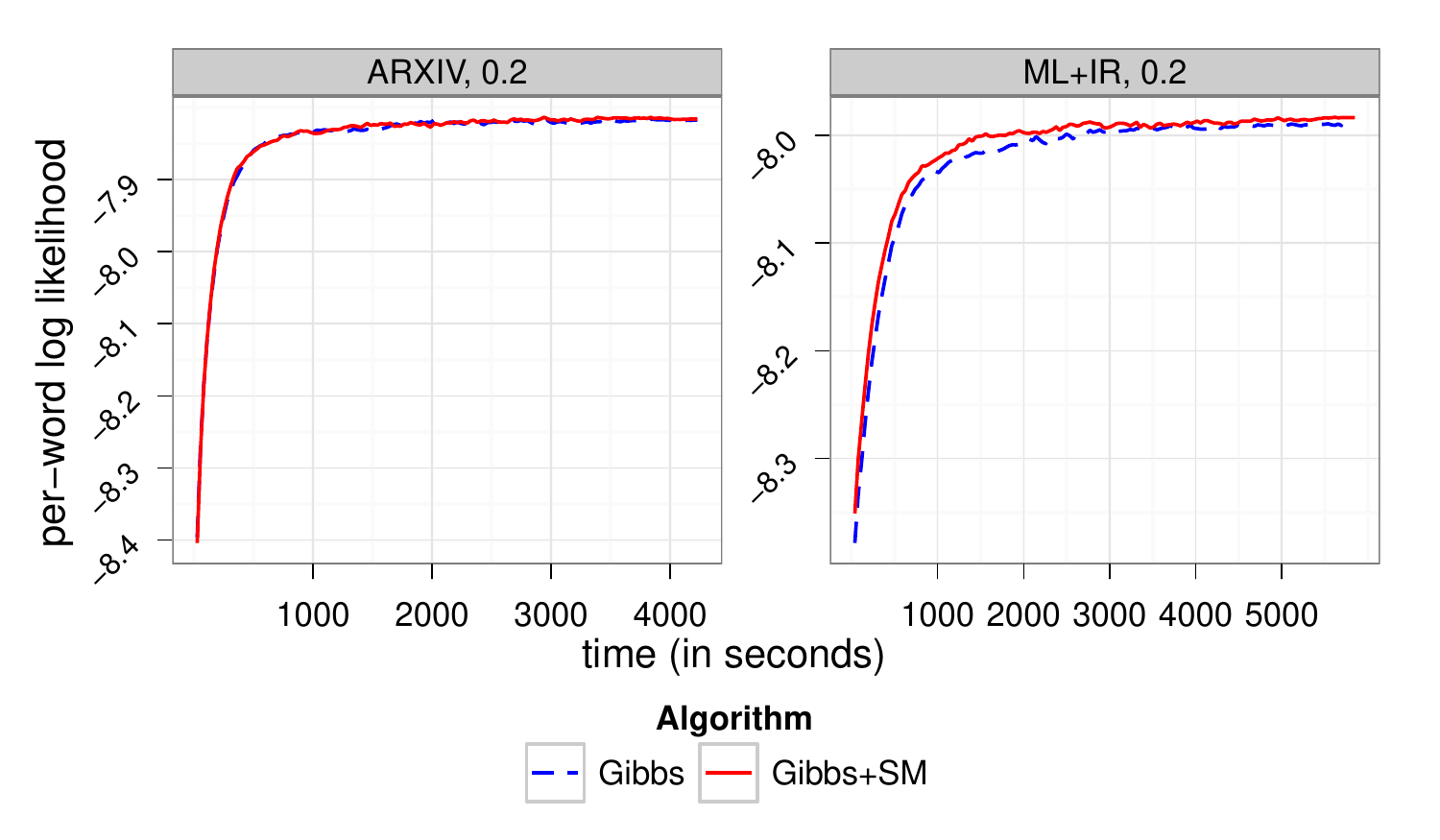}
\caption{The likelihood comparison for ARXIV and ML+IR ($\eta=0.2$). Trends are
similar for other settings.}
\label{fig:likelihood}
\end{figure}

\begin{figure*}
\centering
\includegraphics[width=0.9\textwidth]{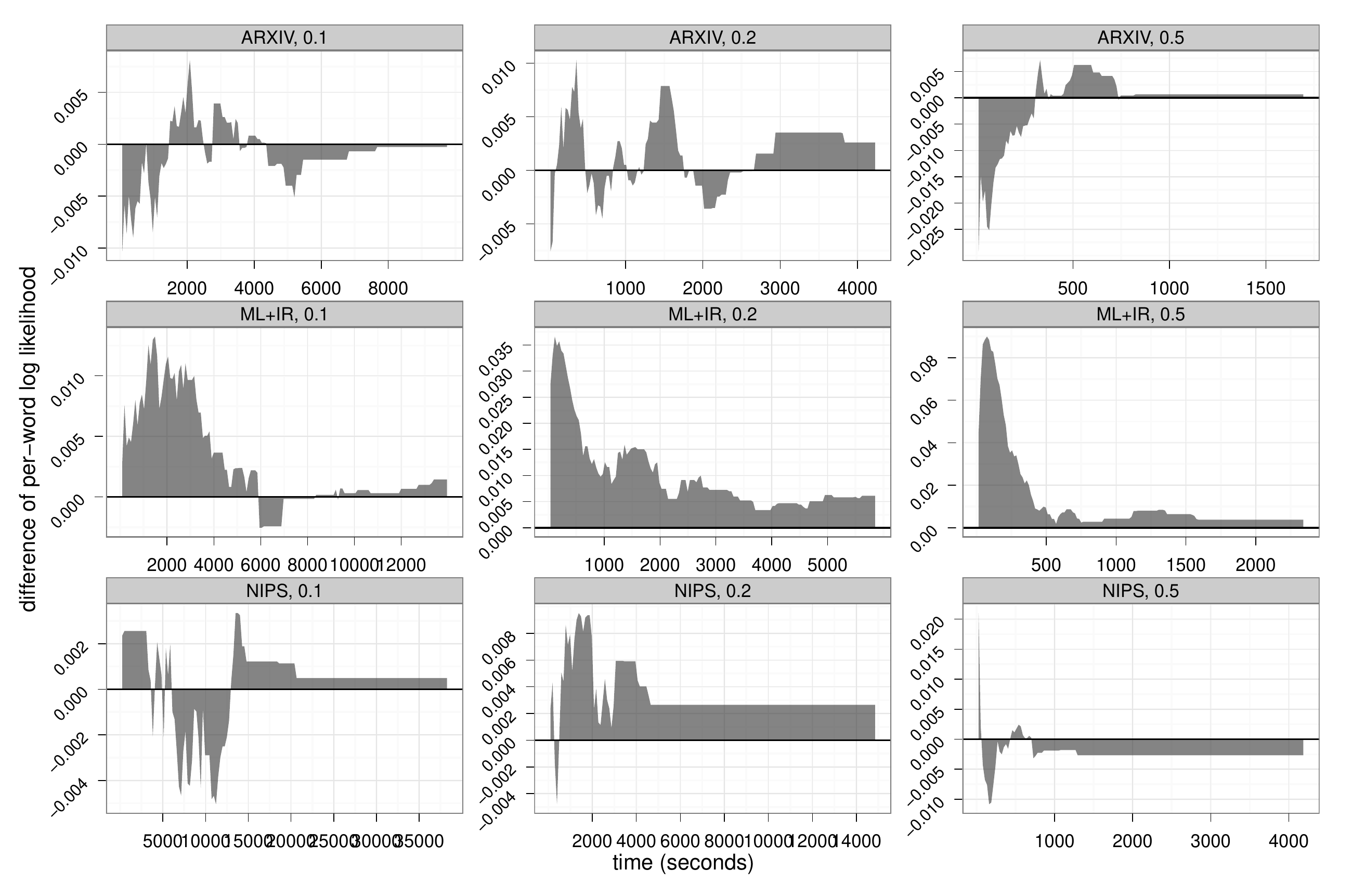}
\caption{Experimental results for the difference of the per-word log likelihood
between Gibbs+SM and Gibbs. Horizontal lines indicate zeros. Results are
averaged over 20 runs, variance is not shown, and time is in log-scale to
allow better view. ``ARXIV, 0.1'' indicates the experiment on ARXIV with
$\eta=0.1$. Others are similarly defined.}
\label{fig:realtext}
\end{figure*}

\begin{figure*}
\centering
\includegraphics[width=0.9\textwidth]{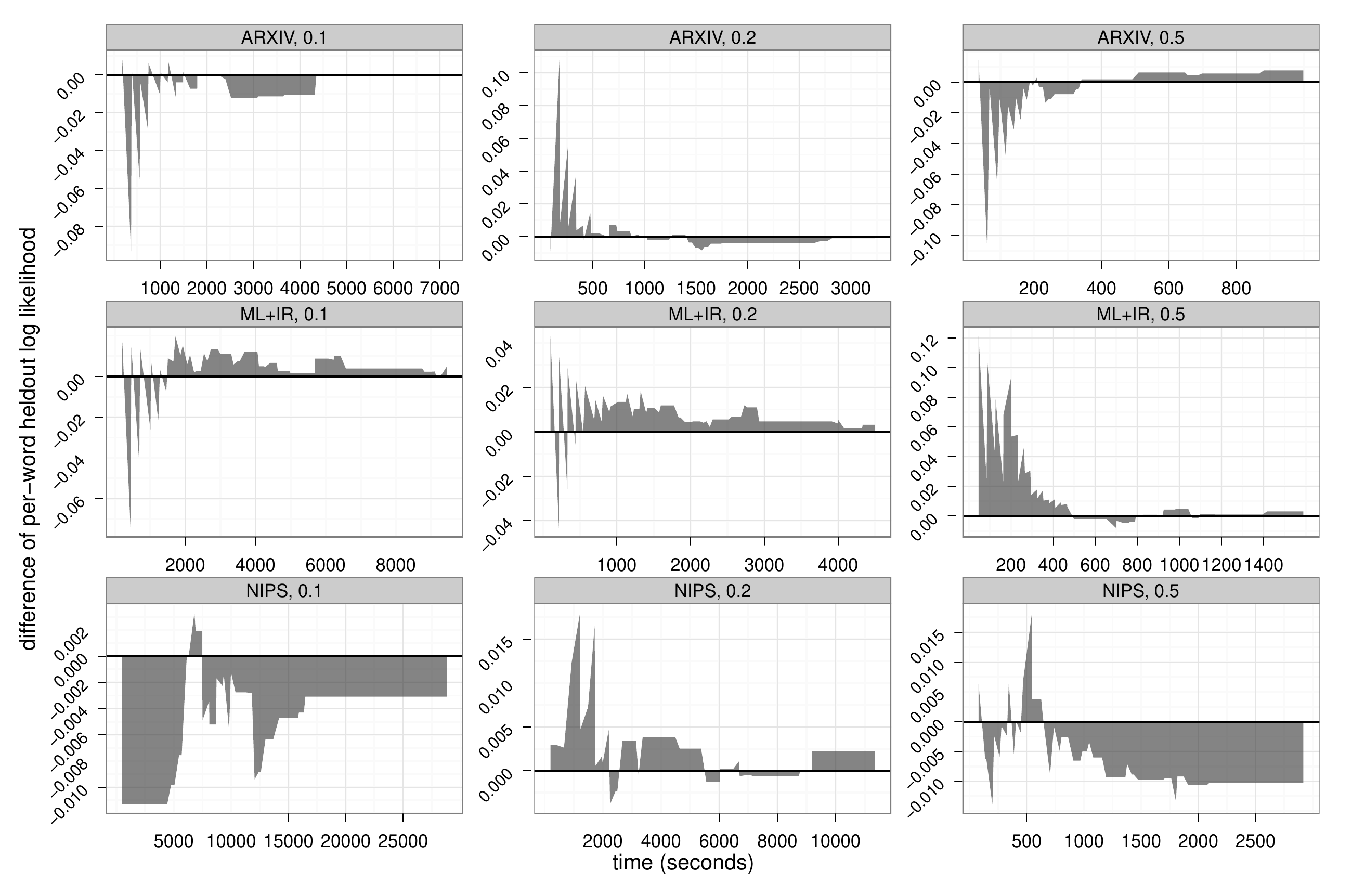}
\caption{Difference of the per-word heldout likelihood on test set between
Gibbs+SM and Gibbs. Horizontal lines indicate zeros. Settings are the
same as those in Figure~\ref{fig:realtext}.}
\label{fig:heldout}
\end{figure*}

HDP analysis is unsupervised, so there is no ground truth. Thus we
compare algorithms by only examining the modes using the per-word log
likelihood of the training set (80\% entire data) and the per-word
heldout log likelihood, for the testing set (20\% entire data). We use
hyperparameters $\gamma$ and $\alpha$ with Gamma priors ${\rm
  Gamma}(1., 1.)$.  We let $\eta=0.1, 0.2, 0.5$.  In general a smaller
$\eta$ leads to more topics, because the prior enforces that the
topics are sparser.  In addition, as we find out in the synthetic
experiments, that it is difficult for split-merge operations make
reverse moves, we only run split-merge operations for the first 50
iterations while the entire Gibbs sampler for 500 iterations.  Since
the data used here is much larger than the synthetic data, we compare
the algorithms using computation time, rather than number of
iterations.  Figures~\ref{fig:realtext} and~\ref{fig:heldout} show the
results. Here we focus on the difference of modes. For the log likelihood
itself, please see example Figure~\ref{fig:likelihood}.


\begin{figure}[t]
\centering
\includegraphics[width=1.0\columnwidth]{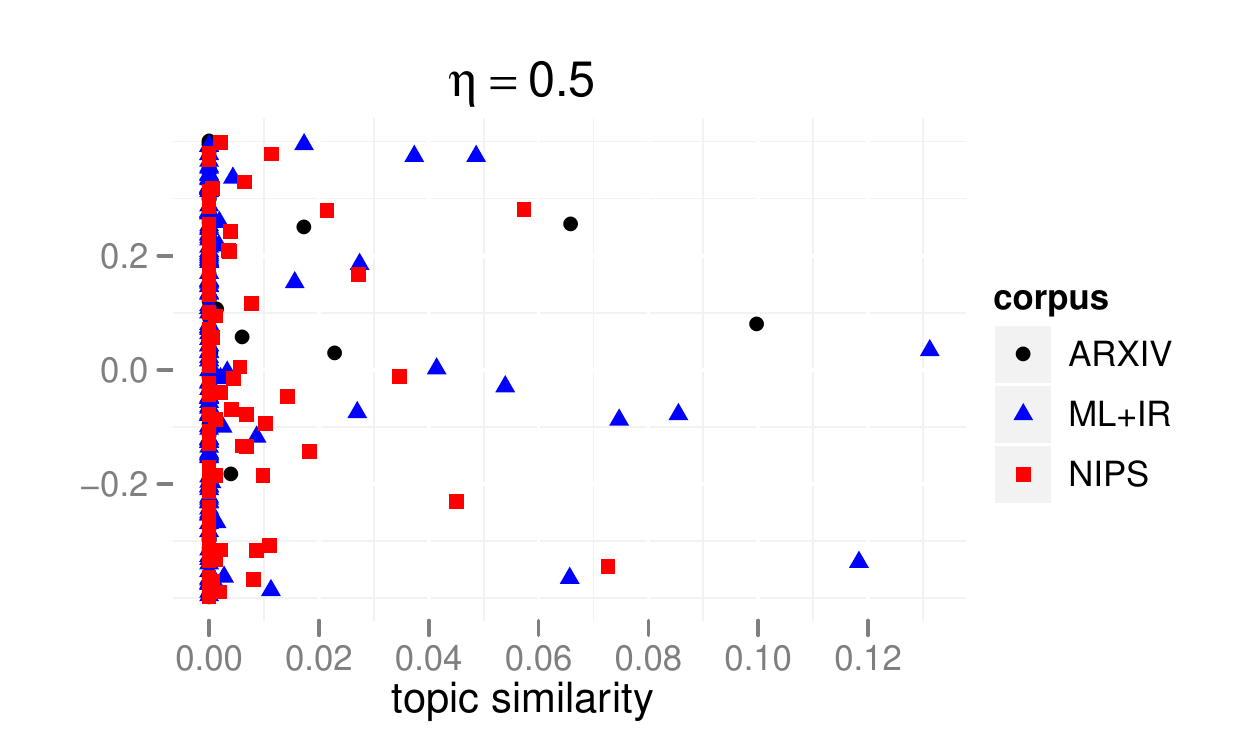}
\caption{Plot of topic similarities.  Each dot's x-axis indicates a
  similarity value of one pair of topics using cosine similarity. (The
  y-axis is for jitter and does not carry meaning.) The ranking of
  having similar topics is, ML+IR $>$ NIPS $>$ ARXIV.}
\label{fig:sim}
\end{figure}

This experimental results are summarized as follows. Towards the end
of the runs, Gibbs+SM and Gibbs are not too different.  Gibbs+SM
reaches a better mode faster with $\eta=0.2$ and $\eta=0.5$ for ML+IR
and $\eta=0.2$ for NIPS, while on other settings, Gibbs+SM was equally
as good as Gibbs sampling.

We hypothesize that there are two reasons.  First, as we see from the
experiments on synthetic data, if topics are easy to separate, the
traditional Gibbs sampler will find them easily as well.  In
Figure~\ref{fig:sim}, we plot the topic similarities in different
corpora for $\eta=0.5$.  We see that ARXIV and NIPS do not have many
similar topics. For ML+IR, however, there are more similar topics.
(This is also demonstrated in~\cite{Wang:2009}.)  Gibbs+SM is more
likely to work well in this scenario.

Second, for $\eta=0.2$ and $\eta=0.5$ in ML+IR, there are fewer topics (but not
few enough so that they cannot be distinguished) than when $\eta=0.1$ in the
corpus---each topic contains more tables, and thus the chance of picking two
informative tables that can serve a good guidance for the split-merge operation
is higher.  For NIPS, the reason it works for $\eta=0.2$ is the same as ML+IR,
for $\eta=0.5$, the corpus might just need very few topics to explain itself and
thus the topics obtained are easy to separate.




In summary, on real data, Gibbs+SM is at least as good as Gibbs
sampling and sometimes helps speed convergence. On average,
split-merge operations are accepted at around $3\%$. In general, the
split-merge operations improve performance when sets of similar topics
exist in the corpus.

\section{CONCLUSIONS AND FUTURE WORK}

We presented a split-merge MCMC algorithm for the HDP topic model.  We
showed on both synthetic and real data that split-merge MCMC algorithm
is effective during the burn-in phase of HDP Gibbs sampling.  Further,
we gave intuitions for what properties of the data lead to improved
performance from split-merge MCMC.

Recently, Gibbs samplers based on the distance dependent Chinese
restaurant process (ddCRP)~\cite{Blei:2010a} have demonstrated
improved convergence for DP mixture models. Applying these ideas to
the HDP is worth exploring.

\section*{Acknowledgement}
Chong Wang is supported by a Google PhD fellowship. David M. Blei is supported
by ONR 175-6343, NSF CAREER 0745520, AFOSR 09NL202, the Alfred P.  Sloan
foundation, and a grant from Google.

\bibliography{journal}
\bibliographystyle{spmpsci}

\section*{Appendix}
We review Gibbs sampling for $\bm t$~\cite{Teh:2007}.  Define the conditional
density of $x_{ji}$ given all words in topic $k$ except $x_{ji}$,
\begin{align*}
    f_k^{-{x}_{ji}}({x}_{ji}) = \frac{\int f(x_{ji}|\phi_k) \prod_{j'i'\neq ji,
    z_{j'i'}=k}f(x_{j'i'}|\phi_k)h(\phi_k){\rm d} \phi_k}{\int \prod_{j'i'\neq ji,
    z_{j'i'}=k}f(x_{j'i'}|\phi_k)h(\phi_k){\rm d} \phi_k},
\end{align*}
$f(\cdot|\phi)$ is the ${\rm Mult}(\phi)$ and $h(\cdot)$ is the density of
$H$, ${\rm Dirichlet}(\eta)$.
Since $h(\phi)$ is a Dirichlet distribution,
$f_k^{-{x}_{ji}}({x}_{ji})$,
\begin{align*}
&f_k^{-{x}_{ji}}({x}_{ji}=v) = \frac{n_{\cdot \cdot k}^{-x_{ji},v}+\eta}{
n_{\cdot \cdot k}^{-x_{ji}}+V\eta},
\end{align*}
Note $f_k^{-{\bm x}_{jt}}({\bm x}_{jt})$ and
$f_k(\{x_{ji}:z_{ji}=k\})$ in the main text can be similarly derived.


 The likelihood for $t_{ji}=t$, when
$t=1, \dots, m_{j\cdot}$ is $f_{k_{jt}}^{-x_{ji}}(x_{ji})$, since
table $t$ is linked to the topics through $k_{jt}$. For $t_{ji}=t_{\rm
  new}$, the likelihood is calculated by integrating out $G_0$,
\begin{align*}
& p(x_{ji}|t_{ji}=t_{\rm new}, {\bm t}^{-ji}, {\bm k}) =\nonumber\\
&  \sum_{k=1}^K\frac{m_{\cdot k}}{m_{\cdot
\cdot}+\gamma}f_k^{-{x}_{ji}}({x}_{ji}) + \frac{\gamma}{m_{\cdot
\cdot}+\gamma}f_{k_{\rm new}}^{-{x}_{ji}}({x}_{ji}),
\end{align*}
where $f_{k_{\rm new}}^{-{x}_{ji}}({x}_{ji})=\int
f(x_{ji}|\phi)h(\phi){\rm d}\phi$ is the prior density for
$x_{ji}$. We have
\begin{align*}
p(&t_{ji} = t \g{\bm t}^{-ji}, {\bm k}) \\
&\propto \left\{ \begin{array}{ll}
    n_{jt\cdot}^{-ji} f_{k_{jt}}^{-x_{ji}}(x_{ji}),& \textrm{if $t_{ji}=1,\dots,
    m_{j\cdot}$},\\ \alpha_0 p(x_{ji}|t_{ji}=t_{\rm new}, {\bm t}^{-ji}, {\bm
    k}), & \textrm{if $t_{ji}=t_{\rm new}$}.
\end{array} \right. \nonumber
\end{align*}
When $t_{ji}=t_{\rm new}$, we need to assign $k_{jt_{\rm new}}$,
\begin{align*}
p(k_{jt_{\rm new}}=k|{\bm t}, {\bm k}^{-jt_{\rm new}}) \propto \left\{ \begin{array}{ll} m_{\cdot k} f_{k}^{-x_{ji}}(x_{ji}),&
\textrm{if $k=1,\dots, K$},\\ \gamma f_{k_{\rm
new}}^{-{x}_{ji}}({x}_{ji}), & \textrm{if $k=k_{\rm new}$}.
\end{array} \right. \nonumber
\end{align*}

\end{document}